\documentclass{article}

\usepackage{arxiv}          
\usepackage[utf8]{inputenc} 
\usepackage[T1]{fontenc}    
\usepackage{geometry}       
\usepackage{hyperref}       
\hypersetup{
    colorlinks=true,
    linkcolor=blue,
    filecolor=magenta,
    urlcolor=cyan,
    pdftitle={Multi-Agent RAG System},
    bookmarks=true,
    pdfpagemode=FullScreen,
}

\usepackage{amsmath, amssymb, amsfonts} 
\usepackage{booktabs}                   
\usepackage{multirow}                   
\usepackage{algpseudocode}

\usepackage{graphicx}       
\graphicspath{{./images/}}  
\usepackage{float}          
\usepackage{caption}        

\usepackage{algorithm}      
\usepackage{algpseudocode}  
\usepackage{listings}       
\usepackage{xcolor}         
\usepackage{microtype}      
\usepackage{nicefrac}       

\usepackage{enumitem}       

\usepackage{hyperref} 
\usepackage{orcidlink} 
\usepackage{graphicx} 

\title{A Collaborative Multi-Agent Approach to Retrieval-Augmented Generation Across Diverse Data Sources}

\author{
 Aniruddha Salve \scalebox{1.2}{\orcidlink{0009-0005-0229-9805}} \\ 
  iASYS Technology Solutions Pvt. Ltd.\\
  Pune, Maharashtra, India \\
  \texttt{aniruddha.salve@iasys.co.in} \\
   \And
 Mahesh Deshmukh \scalebox{1.2}{\orcidlink{0009-0007-4033-5461}}\\
  iASYS Technology Solutions Pvt. Ltd.\\
  Pune, Maharashtra, India \\
  \texttt{mahesh.deshmukh@iasys.co.in} \\
  \And
 Saba Attar \scalebox{1.2}{\orcidlink{0009-0009-9970-4326}}\\
  SVPM's College of Engineering\\
  Baramati, Pune, Maharashtra, India \\
  \texttt{sabaattar1702@gmail.com} \\
  \And
 Sayali Shivpuje \scalebox{1.2}{\orcidlink{0009-0008-1265-8826}}\\
  SVPM's College of Engineering\\
  Baramati, Pune, Maharashtra, India \\
  \texttt{shivpujesayali.2243@gmail.com} \\
  \And
 Arnab Mitra Utsab \scalebox{1.2}{\orcidlink{0009-0000-7351-7842}}\\
  School of Data and Sciences\\ 
  Brac University\\
  Dhaka, Bangladesh \\
  \texttt{arnab.mitra.utsab@g.bracu.ac.bd}
}

\begin{document}
\maketitle



\begin{abstract}
Retrieval-Augmented Generation (RAG) enhances Large Language Models (LLMs) by incorporating external, domain-specific data into the generative process. While LLMs are highly capable, they often rely on static, pre-trained datasets, limiting their ability to integrate dynamic or private data. Traditional RAG systems typically use a single-agent architecture to handle query generation, data retrieval, and response synthesis. However, this approach becomes inefficient when dealing with diverse data sources, such as relational databases, document stores, and graph databases, often leading to performance bottlenecks and reduced accuracy.

This paper proposes a Multi-Agent RAG system to address these limitations. Specialized agents, each optimized for a specific data source, handle query generation for relational, NoSQL, and document-based systems. These agents collaborate within a modular framework, with query execution delegated to an environment designed for compatibility across various database types. This distributed approach enhances query efficiency, reduces token overhead, and improves response accuracy by ensuring that each agent focuses on its specialized task.

The proposed system is scalable and adaptable, making it ideal for generative AI workflows that require integration with diverse, dynamic, or private data sources. By leveraging specialized agents and a modular execution environment, the system provides an efficient and robust solution for handling complex, heterogeneous data environments in generative AI applications.
\end{abstract}


\newpage
\section{Introduction}
Large Language Models (LLMs) have significantly advanced natural language processing by enabling sophisticated query interpretation and text generation. \cite{naveed2024overview, wu2024llm} Despite their capabilities, LLMs are limited by their reliance on static pre-trained datasets, which restricts their ability to incorporate dynamic, domain-specific, or private data into their responses. Retrieval-Augmented Generation (RAG) systems address this challenge by integrating external data retrieval with generative processes, providing more context-aware and accurate outputs.

Traditional RAG systems typically employ single-agent architectures where a single system is responsible for query generation, data retrieval, and response synthesis. While effective for basic use cases, these monolithic designs often face limitations when dealing with diverse data sources, such as relational databases, document stores, and graph-based data \cite{liang2024aligning}. These systems also require elaborate prompts containing schemas, examples, and user queries, leading to inefficiencies in token usage, increased processing latency, and potential inaccuracies in query handling.

To overcome these challenges, this paper proposes a Multi-Agent RAG system \cite{talebirad2023multiagent,guo2024multiagent}. Unlike traditional approaches, this system delegates the task of query generation to specialized agents, each tailored to a specific type of database. These agents generate optimized, database-specific queries without directly executing or retrieving data. Queries are executed in a separate execution environment, ensuring compatibility with diverse data storage systems. The retrieved context is then combined with the user's original query and processed by a generative agent, which synthesizes a coherent and contextually relevant response.

The proposed system introduces a modular design that improves scalability, enhances retrieval accuracy, and supports seamless integration with polyglot data environments \cite{glake2022polyglot}. By delegating specific tasks to specialized agents and incorporating a centralized execution mechanism, this system addresses the inefficiencies of traditional single-agent \cite{sumers2024cognitive} RAG implementations. It offers a robust solution for integrating domain-specific and heterogeneous data into generative AI workflows, making it well-suited for a wide range of real-world applications.

\section{Background}
\subsection{Retrieval-Augmented Generation (RAG)}
Retrieval-Augmented Generation (RAG) extends the capabilities of LLMs by incorporating external or private data into the response generation process. Unlike traditional LLMs that generate responses based solely on their pre-trained knowledge \cite{naveed2024overview}, RAG systems retrieve relevant domain-specific data from external sources to provide contextually accurate and up-to-date outputs. This is particularly useful when dealing with private or proprietary data that is inaccessible to the LLM during training. \cite{lewis2020retrieval, chaubey2024comparative, ziletti2024retrieval}

The retrieval phase is critical in RAG as it forms the foundation for accurate and relevant responses. For example, when a user asks, “What is Machine Learning?” the LLM can generate a generic response from its pre-trained knowledge. However, if the user asks, “What is the role of AI and ML in the XYZ development team?” the system needs to retrieve private context, such as “The XYZ team uses AI for predictive analytics and natural language processing,” to provide a tailored and meaningful response. The integration of retrieved context with generative capabilities ensures that the outputs are both accurate and relevant.

Traditional RAG Systems, which rely on single-agent architectures, face challenges in managing multiple data sources. These systems are prone to inefficiencies, including token overload and inaccurate query generation, especially when handling diverse data storage types. The need for a more scalable and efficient approach has led to the development of the Multi-Agent RAG system. 




\subsection{Agents and Multi-Agent RAG Systems}

In a Retrieval-Augmented Generation (RAG) \cite{lewis2020retrieval} system, a single agent \cite{sumers2024cognitive} is capable of handling query generation and data retrieval for multiple types of databases, including relational, document-based, and graph data sources \cite{liang2024aligning}. However, this approach demands that the agent be equipped with a comprehensive schema, detailed examples, and extensive knowledge of diverse database structures. This complexity can lead to inefficiencies such as slower query generation, token overhead, and a higher likelihood of errors in query formation. Additionally, relying on a single agent to process heterogeneous data increases the risk of misinterpretation, resulting in incorrect or suboptimal queries. These limitations often hinder the scalability and performance of systems operating in diverse data environments.

The Multi-Agent \cite{talebirad2023multiagent,guo2024multiagent} RAG system resolves these challenges by delegating tasks to specialized agents, each tailored to handle a specific type of database. For example, a MySQL agent is responsible for generating optimized SQL queries for relational databases \cite{ziletti2024retrieval}, while a MongoDB agent specializes in document-based data retrieval. By focusing solely on one database type, each agent can generate precise and efficient queries, reducing the workload and complexity associated with a single-agent architecture. This division of labor not only improves accuracy and speed but also allows the system to scale seamlessly by adding new agents for emerging data sources. The modular design ensures adaptability, making the Multi-Agent \cite{talebirad2023multiagent,guo2024multiagent} RAG system robust and effective in managing diverse data environments.

\subsection{Choice of LLM in RAG}

The choice of Large Language Models (LLMs) \cite{naveed2024overview, yan2024local_llms, bumgardner2023local, qu2024compact} is a crucial consideration in the design and implementation of Retrieval-Augmented Generation (RAG) systems. LLMs serve as the foundation for query generation and response synthesis, but their token limits and processing capabilities significantly impact the system's efficiency and scalability. LLMs can be broadly categorized into two types: local models \cite{an2024local_llms} and API-based models \cite{openai2024gpt_models, gemini2024models, yan2024local_llms}. Both categories have unique advantages and challenges, making the decision highly dependent on factors such as data sensitivity, computational resources, and scalability requirements.

Local LLMs, such as Mistral, Zephyr, and Llama \cite{bumgardner2023local}, operate within an organization’s infrastructure, providing complete control over sensitive or proprietary data. This makes them particularly suitable for industries like healthcare and finance, where data privacy and regulatory compliance are critical. Local deployment ensures that no sensitive data leaves the premises, mitigating risks associated with data breaches. However, these models require substantial computational resources, including high-performance GPUs, for efficient operation, which can increase the cost and complexity of deployment.\cite{yan2024local_llms, qu2024compact}.

API-based LLMs, such as OpenAI’s GPT models and Google’s Gemini family, are hosted on cloud platforms and eliminate the need for maintaining extensive infrastructure. They are easy to integrate, benefit from continuous updates, and provide high scalability, making them ideal for applications requiring dynamic workloads. Despite these advantages, API-based models introduce concerns regarding data privacy, as sensitive information must be transmitted to external servers for processing. Additionally, their token-based pricing models can lead to unpredictable costs for applications with heavy usage. \cite{openai2024gpt_models, gemini2024models}

Token constraints further influence the choice of LLMs in RAG systems. Both local and API-based models impose limits on the amount of data that can be processed in a single query, necessitating the use of efficient data chunking and retrieval techniques. Table~\ref{tab:llm_comparison} summarizes the key characteristics of various local and API-based LLMs, focusing on their context lengths and providers.


\begin{table}[h!]
\centering
\begin{tabular}{lllr}
\toprule
\textbf{Model Type}         & \textbf{LLM Provider} & \textbf{LLM Model}         & \textbf{Context Window} \\ 
\midrule
\multirow{4}{*}{Local LLMs} & Mistral               & Mistral 7B                & 4096 - 16K (Sliding Windows) \\ 
                             & Huggingface          & Zephyr                    & 8192 \\ 
                             & Microsoft            & Phi-2                     & 2048 \\ 
                             & Meta                 & Llama 3                   & 8192 \\ 
\midrule
\multirow{5}{*}{API-Based LLMs} & OpenAI            & GPT-4                     & 128,000 \\ 
                                & OpenAI            & GPT-4 Turbo               & 128,000 \\ 
                                & OpenAI            & GPT-3.5 Turbo             & 16,385 \\ 
                                & Google            & Gemini 1.5 Flash          & 1,048,576 \\ 
                                & Google            & Gemini 1.5 Pro            & 2,097,152 \\ 
\bottomrule
\end{tabular}
\caption{Comparison of Local and API-Based LLMs.}
\label{tab:llm_comparison}
\end{table}

The selection of an LLM for RAG systems depends on a balance of priorities. Local models are better suited for scenarios where data privacy and control are paramount, while API-based models excel in environments that demand scalability and ease of integration. Hybrid approaches that combine local processing for sensitive data with cloud-based solutions for less critical tasks represent a promising direction for future advancements. Such system could offer the best of both worlds, enabling organizations to achieve a balance between privacy, performance, and cost-effectiveness in RAG systems.

\subsection{Prompt Engineering for Multi-Agent RAG Systems}
Prompt engineering is a crucial aspect of the Multi-Agent RAG System, as it directly influences the quality and accuracy of the generated queries \cite{sahoo2024prompt, trad2024prompt, yang2023improve}. Few-shot prompting is particularly effective in this context, as it provides the agent with examples of how to interact with a specific database schema. A well-designed prompt includes the user’s query, the database schema, and a few-shot example illustrating how similar queries have been constructed.\cite{trad2024prompt}

Few-shot prompts are carefully crafted to minimize errors and optimize token usage. For example, a MongoDB agent tasked with retrieving research papers might be given a schema detailing the structure of the database and examples of how to construct queries for different use cases. This targeted approach ensures that agents generate accurate and efficient queries tailored to their specific data domains.


\paragraph{ElasticSearch Agent Few-Shot Prompt:} 
\subparagraph{Schema Overview:}
\begin{verbatim}
{
  "index": "SupportTickets",
  "fields": {
    "ticket_id": "string",
    "description": "string",
    "raised_by": "string",
    "status": "string"
  }
}
\end{verbatim}

\subparagraph{Example 1:}

\subparagraph{Question:} Find support tickets related to MySQL issues raised by Sayali Shivpuje. 
\textbf{\\ Expected Query}:
\begin{verbatim}
{
  "query": {
    "bool": {
      "must": [
        { "match": { "description": "MySQL" }},
        { "match": { "raised_by": "Sayali Shivpuje" }}
      ]
    }
  }
}
\end{verbatim}

\subparagraph{Example 2:}
\subparagraph{Question:} Retrieve all open tickets related to Neo4j raised by Aniruddha Salve. 
\textbf{\\ Expected Query}
\begin{verbatim}
{
  "query": {
    "bool": {
      "must": [
        { "match": { "description": "Neo4j" }},
        { "match": { "raised_by": "Aniruddha Salve" }},
        { "match": { "status": "open" }}
      ]
    }
  }
}
\end{verbatim}

\paragraph{MySQL Agent Few-Shot Prompt: } 

\subparagraph{Schema Overview:}
\begin{verbatim}
Table: Projects
Columns:
  - project_id (int)
  - project_name (string)
  - assigned_to (string)
  - start_date (date)
  - end_date (date)
  - status (string)
\end{verbatim}

\subparagraph{Example 1: }

\subparagraph{Question:} List all active projects handled by Saba Attar.
\textbf{\\ Expected Query}:

\begin{verbatim}
SELECT project_name 
FROM Projects 
WHERE assigned_to = 'Saba Attar' AND status = 'active';
\end{verbatim}

\subparagraph{Example 2:}
\subparagraph{Question:}
Retrieve all completed projects assigned to Mahesh Deshmukh.
\textbf{\\ Expected Query}:
\begin{verbatim}
SELECT project_name 
FROM Projects 
WHERE assigned_to = 'Mahesh Deshmukh' AND status = 'completed';
\end{verbatim}

\paragraph{MongoDB Agent Few-Shot Prompt}

\subparagraph{Schema Overview:}
\begin{verbatim}
{
  "collection": "Projects",
  "fields": {
    "project_id": "int",
    "assigned_to": "string",
    "status": "string",
    "deadline": "date"
  }
}
\end{verbatim}

\subparagraph{Example 1: }

\subparagraph{Question:} 
Find all active projects assigned to Aniruddha Salve.
\textbf{\\ Generated Query:}
\begin{verbatim}
db.Projects.find({ "assigned_to": "Aniruddha Salve", "status": "active" });
\end{verbatim}

\subparagraph{Example 2:}
\subparagraph{Question:} 
Retrieve all completed projects assigned to Saba Attar.
\textbf{\\ Generated Query:}
\begin{verbatim}
db.Projects.find({ "assigned_to": "Saba Attar", "status": "completed" });
\end{verbatim}

\paragraph{Neo4j Agent Few-Shot Prompt}
\subparagraph{Schema Overview:}
\begin{verbatim}
Graph: ResearchNetwork
Nodes:
  - Researcher {name: string, field: string}
Relationships:
  - COLLABORATES_WITH
  - WORKS_ON
\end{verbatim}
\subparagraph{Example 1:}
\subparagraph{Question:} List all collaborators of Arnab Mitra Utsab.
\textbf{\\ Generated Query:}
\begin{verbatim}
MATCH (r:Researcher {name: 'Arnab Mitra Utsab'})-[:COLLABORATES_WITH]->(collaborator:Researcher)
RETURN collaborator.name;
\end{verbatim}

\subparagraph{Example 2:}
\subparagraph{Question:} Find researchers working on AI projects in the domain of healthcare.
\textbf{\\ Generated Query:}
\begin{verbatim}
MATCH (r:Researcher)-[:WORKS_ON]->(project)
WHERE project.domain = 'AI in Healthcare'
RETURN r.name;
\end{verbatim}

\section{Related Work and Research Contributions}

\subsection{Single-Agent RAG Systems}
This study focuses on improving Single-Agent \cite{sumers2024cognitive} Retrieval-Augmented Generation (RAG) systems to solve complex problems more effectively. It addresses the limitations of traditional LLMs, such as weak reasoning and context handling, by using evidence-based reasoning to guide response generation. This approach simplifies the process by having a single agent retrieve and use relevant evidence for better output. The study shows that single-agent RAG systems can achieve strong performance in knowledge-intensive tasks, offering a more efficient alternative to multi-agent systems while still delivering high-quality results \cite{lewis2020retrieval}.

The Speculative RAG framework enhances single-agent RAG systems by using a two-LM approach for improved efficiency and accuracy. A smaller, distilled specialist LM drafts multiple responses from different document subsets, while a larger generalist LM verifies these drafts in a single pass. This method reduces token use, mitigates position bias, and speeds up processing. Extensive testing on benchmarks such as TriviaQA, MuSiQue, PubHealth, and ARC-Challenge shows that Speculative RAG boosts accuracy by up to 12.97\% and cuts latency by 51\% compared to conventional RAG systems, setting a new standard for single-agent RAG performance \cite{wang2024speculative}.

Although pre-trained language models are good at storing knowledge, they have trouble updating and precisely accessing it, particularly when performing knowledge-intensive operations. To overcome this, RAG models integrate non-parametric memory, such as a vector index of Wikipedia, that is retrieved by a neural retriever with pre-trained parametric memory. This method enhances the model's capacity to produce precise and accurate answers. There are two primary RAG approaches: one modifies the retrieved passages per token, while the other uses the same passages throughout the output. Optimized RAG models have demonstrated improved generation quality and factual correctness on open-domain QA tasks, outperforming task-specific architectures and conventional parametric models \cite{lewis2020retrieval}.

\subsection{Multi-Agent Systems in AI}
By focusing on relevant feedback and multi-agent communication patterns, this work examines how large language models (LLMs) might improve Retrieval-Augmented Generation (RAG) systems for technical support. Larger models outperform traditional approaches in determining relevance, according to experiments with LLMs such as GPT-4, GPT-3.5-turbo, and Llama3. However, smaller models, such as Llama3, produced comparable findings when given appropriate prompts. Although they required more computing power, patterns like Reflection and Planning improved accuracy for multi-agent communication by 55\% compared to simpler techniques. This study demonstrates how LLMs enhance RAG systems while balancing quality and efficiency \cite{finsas2024optimizing}.

The study also examines how to use Large Language Models (LLMs) to enhance 6G communication systems while addressing limitations in logic, data protection, and refinement. Three elements of a proposed multi-agent system improve LLMs: multi-agent evaluation and reflection (MER) to evaluate and improve results; multi-agent data retrieval (MDR) to increase knowledge; and multi-agent collaborative planning (MCP) to provide solutions. This method demonstrates how multi-agent collaborations can enhance LLM performance for 6G tasks, validated through a case study on semantic communication \cite{jiang2024multiagent}.

Furthermore, the Extended Coevolutionary (EC) Theory combines coevolutionary dynamics, adaptive learning, and LLM-based strategy suggestions to overcome the shortcomings of conventional game theory in modeling dynamic Multi-Agent Systems (MASs). Unlike static models, this paradigm accounts for diverse interactions, including social relationships, financial transfers, and information exchange. Simulations validate the EC framework's capacity to promote social welfare, strengthen resilience against shocks, and encourage cooperation in MASs. This method addresses existing issues in MASs and opens the door for further research in adaptive learning and strategic decision-making \cite{yan2022boundary}.

The potential of MASs to optimize complex tasks through cooperation is also discussed, emphasizing privacy protection and cooperative optimization, including distributed and federated approaches. The survey covers cooperative and non-cooperative games, analyzing individual cost reduction in non-cooperative settings and global cost minimization in cooperative ones. Future research prospects in cooperative optimization and game theory applications are highlighted \cite{wang2022cooperative}.

\subsection{Literature Gap Discussion}

RAG systems have advanced significantly, but there are still issues with single-agent and multi-agent techniques requiring further research. Table \ref{tab:literature_gap} outlines key gaps in the current literature.

\begin{table}[h!]
    \centering
    \begin{tabular}{lp{6cm}p{6cm}}
        \toprule
        \textbf{Aspect} & \textbf{Single-Agent RAG} & \textbf{Multi-Agent RAG} \\ 
        \midrule
        Token Efficiency & Improved by frameworks like Speculative RAG but limited by response generation length and context capacity. & Decentralization helps reduce token usage, but coordination across agents can increase token overhead. \\ 
        \midrule
        Contextual Handling & Strong in evidence-based reasoning but struggles with dynamic context changes and context maintenance. & Benefits from specialized agents managing different database types, but integration of diverse agents and data sources remains complex. \\ 
        \midrule
        Scalability & Effective for focused, smaller tasks but not well-suited for large-scale database integration. & Enhanced by distributing tasks, but managing complex interactions and synchronization between agents is challenging. \\ 
        \midrule
        Knowledge Access & Leverages pre-trained and non-parametric memory for improved accuracy but has limitations in real-time knowledge updating. & Multi-agent collaboration helps with knowledge expansion and validation, yet adapting agents to new or changing data types is an open issue. \\ 
        \midrule
        Diversity in Data Sources & Primarily uses indexed data like Wikipedia and suffers when addressing varied, heterogeneous databases. & Can include different database types but maintaining coherence and uniformity across data sources is difficult. \\ 
        \midrule
        Coordination and Communication & Typically involves a single-agent model, limiting interaction complexity. & Multi-agent approaches offer improved handling but require intricate communication strategies, leading to potential inefficiencies. \\ 
        \midrule
        Application Versatility & Effective for specific, narrow tasks but less versatile for complex or varied applications. & Versatile across different database environments, but adapting multi-agent interactions for consistent performance across domains needs more research. \\ 
        \bottomrule
    \end{tabular}
    \caption{Literature Gap Analysis}
    \label{tab:literature_gap}
\end{table}

\subsection{Contribution}

Our work introduces a novel way to improve the accuracy, scalability, and efficiency of database interactions using a multi-agent system for retrieval-augmented generation. By addressing common challenges like token limitations, diverse database types, and inefficient query handling, we aim to rethink how generative AI interacts with databases. Below are the key contributions of our approach:

\subsubsection{Specialized Agents for Diverse Databases}

We designed specialized agents, such as the MySQL Agent, Neo4j Agent, and MongoDB Agent, tailored for specific database types. Each agent is optimized to interact seamlessly with its corresponding database, ensuring accurate queries and efficient data retrieval for relational, graph, and document-oriented systems.

\subsubsection{Centralized Query Execution}

Our system simplifies query execution by combining outputs from different agents in a unified environment. This avoids redundant transformations and streamlines data retrieval across multiple database formats, ensuring that responses are both efficient and context-aware.

\subsubsection{Collaborative Agent Workflows}

Each agent is tasked with specific roles, such as query generation, retrieval, or validation, depending on the database type. By allowing agents to focus on their strengths, we ensure that the generated results are precise and relevant. These outputs are then combined into a cohesive response, enhancing both retrieval quality and generative outputs.

\subsubsection{Integrating Generative AI Thoughtfully}

The data retrieved by agents is synthesized into a coherent response using a central language model. By ensuring that only the most relevant and concise inputs are sent to the generative model, we improve the quality of outputs while reducing unnecessary or irrelevant information.

\subsubsection{Synthesizing Data for Rich Responses}

Once data is retrieved, a generative component creates detailed and meaningful responses by merging information from multiple sources. This step ensures that user queries are addressed comprehensively, resulting in responses that are both logical and contextually rich.

\subsubsection{Error Handling for Robust Performance}

Our system includes mechanisms to detect and address errors during query execution. When an issue arises, it automatically switches to fallback methods to ensure that users receive reliable results, even in challenging scenarios.

\subsubsection{Reducing Token Overhead}

By distributing tasks among specialized agents, we limit the token consumption for individual queries. This modular approach allows the system to handle more complex tasks within predefined token constraints, improving its scalability.

\subsubsection{Adaptable Across Industries}

The system can adapt to different database types and environments without requiring major changes. Its flexibility makes it suitable for applications in fields like healthcare, finance, and logistics, where seamless integration with diverse data sources is essential.

\subsubsection{Efficient Use of Resources}

By distributing workloads among agents and using a single generative model for synthesis, we significantly reduce computational requirements. This efficiency is particularly beneficial in real-time applications or environments with limited processing power.

\subsubsection{Improving Query Accuracy}

Each agent’s expertise ensures that database queries are tailored and handled with high precision. By collaborating, the agents reduce context loss, resulting in more accurate and meaningful responses.

\subsubsection{Enabling Future Research in Multi-Agent Systems}

This work highlights the potential of multi-agent collaboration for enhancing retrieval-augmented generation. By exploring agent specialization and synthesis, we provide a stepping stone for further advancements in this area.

Through these contributions, we address the limitations of existing approaches and demonstrate how intelligent, specialized systems can enable more efficient and accurate interactions between generative AI and complex data environments.

\section{Proposed System Overview}
The proposed Multi-Agent Retrieval-Augmented Generation (RAG) system utilizes a modular and scalable architecture to integrate multiple data sources with generative AI capabilities. Traditional single-agent RAG systems, while effective in simpler use cases, struggle with complex queries across various data environments. These systems typically rely on one entity that handles query generation, data retrieval, and response synthesis. This approach often results in inefficiencies when dealing with heterogeneous data sources such as relational, document-based, and graph-based data, leading to performance bottlenecks and inaccuracies in query processing.

To address these challenges, the proposed system decentralizes task management by introducing specialized agents, each optimized for handling different types of data. For instance, the relational agent processes queries for relational data sources, while the graph agent manages graph data, the document agent works with document-based storage, and the text agent handles text-based data sources. These specialized agents collaborate within a centralized query execution environment, ensuring seamless compatibility across diverse data types. Once the relevant data is retrieved, it is passed to the generative agent, which synthesizes it into a coherent and contextually relevant response. The system's modular approach enhances the overall query efficiency, reduces token overhead, and improves the accuracy of the final output, making it suitable for real-world applications that involve complex, multi-source data environments.
\begin{figure}[H]
    \centering
    \includegraphics[width=0.8\textwidth]{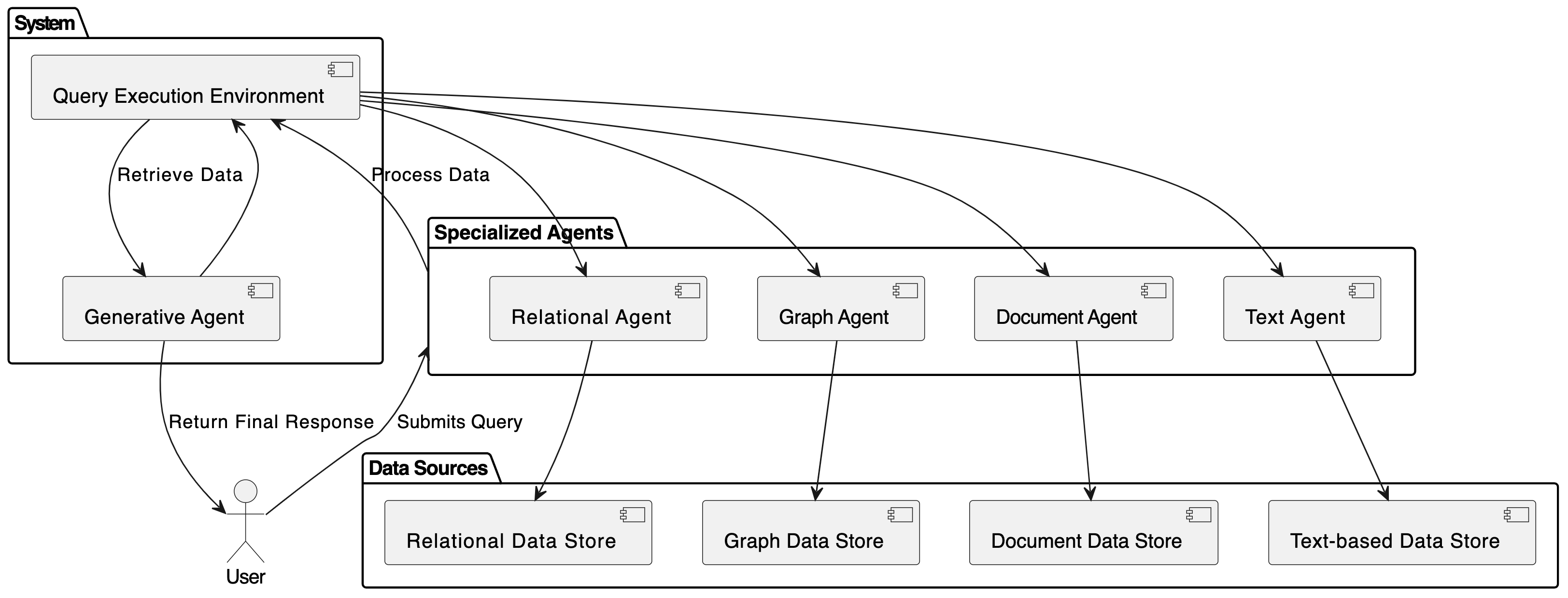} 
    \caption{Proposed Multi-Agent RAG System }
    \label{fig:system-architecture}
\end{figure}

\section{Architecture Overview}

The proposed system introduces a modular and efficient architecture that leverages specialized agents to handle data retrieval and response generation. This architecture ensures seamless integration and processing across various data types and sources while maintaining scalability and adaptability. The key components of the architecture include query generation agents, the query execution environment, the generative agent, and agent specialization. These components work in harmony to achieve efficient query generation, execution, and synthesis of accurate responses.

\subsection{Components of the Architecture}

\subsubsection{Query Generation Agents}
Query generation agents are essential to the proposed system, as they are responsible for generating executable queries that are tailored to the specific data sources. These agents operate independently, with each one specializing in handling a particular type of database. For example, the MySQL agent generates SQL queries optimized for relational databases, ensuring that the structure and relationships between tables are correctly represented. The MongoDB agent, on the other hand, formulates queries suited for document-based NoSQL databases, focusing on handling nested and hierarchical data. Similarly, the Neo4j agent is specialized in generating Cypher queries, which are used for graph-based databases, enabling efficient querying of relationships between entities.

Each agent utilizes the user’s query and the database schema to create the most effective query for that specific data source. This process is guided by a set of few-shot prompts \cite{trad2024prompt} that provide the agent with relevant examples, ensuring that the query is syntactically correct and semantically meaningful. The modular approach to query generation allows the system to adapt to a wide range of data environments, making it suitable for complex, real-world applications. The query generated by each agent, \( Q_{\text{generated}} \), is derived from the user’s query, \( Q_{\text{user}} \), and the database schema, \( S_{\text{schema}} \), through a function \( f_{\text{agent}} \), as expressed by the equation:

\[
Q_{\text{generated}} = f_{\text{agent}}(Q_{\text{user}}, S_{\text{schema}})
\]

This ensures that the generated query is perfectly aligned with the target database, which allows it to retrieve the correct data as requested by the user.

\subsubsection{Query Execution Environment}
After the query is generated by the appropriate query generation agent, it is passed to the \textbf{Query Execution Environment}. This environment serves as a central platform that connects the generated queries with the corresponding databases to retrieve the necessary data. The query execution environment plays a crucial role in ensuring that the queries are executed correctly and efficiently, managing communication between the system and different types of databases.

The environment is equipped with database drivers that ensure compatibility with various types of databases. For example, relational databases like MySQL are accessed through JDBC drivers, which handle structured data in tables. For NoSQL databases such as MongoDB, there are MongoDB-specific drivers that handle document-based data, which can be more flexible and hierarchical. Similarly, graph databases like Neo4j require Neo4j drivers, which are specifically designed to query relationships between nodes in a graph structure. Depending on the type of database being queried, the correct driver is chosen to ensure smooth communication and efficient data retrieval.

The query execution environment is responsible for managing the database connections, executing the queries generated by the agents, and retrieving the results. The execution function, \( g_{\text{db}} \), handles the query \( Q_{\text{generated}} \) and interacts with the target database \( D_{\text{connection}} \) to return the results \( R_{\text{query}} \), which are then passed back to the system for further processing. This relationship is captured in the equation:

\[
R_{\text{query}} = g_{\text{db}}(Q_{\text{generated}}, D_{\text{connection}})
\]

By managing query execution in this way, the environment ensures that the system can efficiently retrieve data from various database types while maintaining consistency and accuracy across different data sources.\cite{glake2022polyglot}

\subsubsection{Generative Agent}
The \textbf{Generative Agent} is responsible for synthesizing the final response to the user’s query based on the retrieved data. Once the query has been executed and the relevant data has been retrieved, the generative agent processes this information to create a coherent and contextually relevant response. This component serves as the "creative engine" of the system, transforming raw data into a format that is both understandable and useful for the user.

The generative agent uses both the original user query, \( Q_{\text{user}} \), and the retrieved data, \( R_{\text{query}} \), to produce the final output, \( A_{\text{response}} \). The response could take many forms, depending on the user’s request, such as structured text, tables, or graphs. The equation describing the relationship between the user query, the retrieved data, and the final response is:

\[
A_{\text{response}} = h_{\text{gen}}(Q_{\text{user}}, R_{\text{query}})
\]

This ensures that the response is not only accurate but also contextually relevant, addressing the user’s needs based on the data retrieved from various sources.

\subsubsection{Agent Specialization}
Agent specialization is a core feature of the proposed system, as each agent is specifically designed to handle a certain type of data or database. This specialization allows the system to efficiently process queries with a high degree of accuracy and performance. For instance, the MySQL agent is specialized for relational data, making it well-suited for handling queries involving structured databases with complex relationships. The Neo4j agent excels at handling graph data, efficiently managing relationships between entities in graph-based databases. Similarly, the MongoDB agent is optimized for document-oriented data, making it ideal for working with flexible, hierarchical data structures common in NoSQL systems.

The system’s modular architecture ensures that agents can be added or updated without disrupting the overall functionality. As new data types or sources emerge, new agents can be integrated to support these technologies. This flexibility is key to ensuring the system’s long-term scalability and adaptability. The function \( f_{\text{specialization}} \) captures the relationship between the type of data, \( D_{\text{data type}} \), and the agent type, \( A_{\text{agent type}} \), as follows:

\[
S_{\text{agent}} = f_{\text{specialization}}(D_{\text{data type}}, A_{\text{agent type}})
\]

This approach ensures that each agent is tailored for the specific requirements of the data it handles, allowing the system to deliver precise and optimized query results across a variety of data environments.

\subsection{Architecture Diagram}
The architecture diagram illustrates the interconnected components of the system, showcasing how query generation agents, the query execution environment, and the generative agent work together to deliver accurate and contextually relevant responses. It highlights the modular nature of the system, emphasizing the role of each specialized agent in handling specific data types and the seamless integration of these components to achieve efficient query generation, execution, and response synthesis.

\begin{figure}[H]
    \centering
    \includegraphics[scale=0.5]{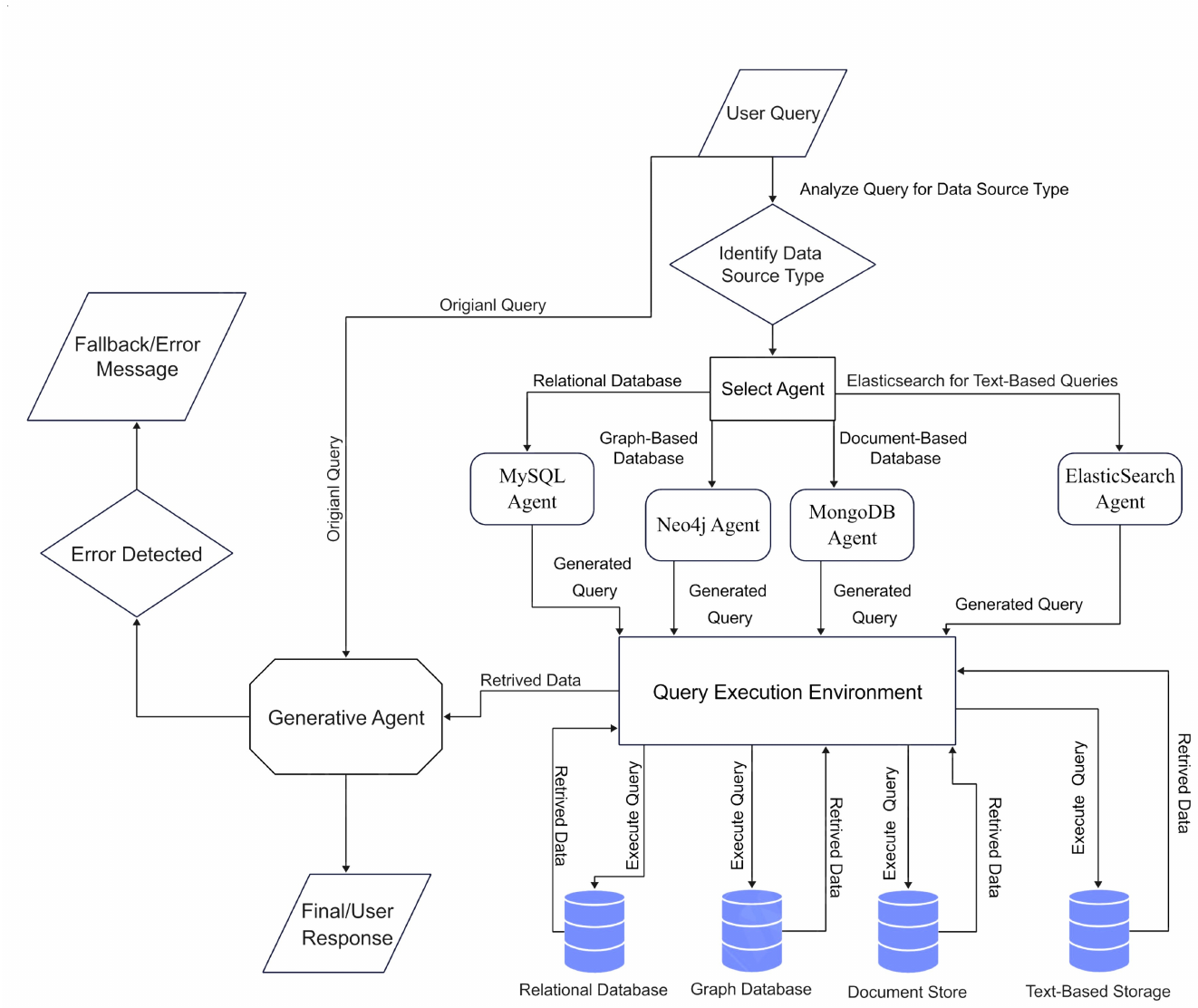}
    \caption{Multi-Agent RAG System Architecture}
    \label{fig:architecture}
\end{figure}


This architecture ensures that the proposed system remains efficient and scalable while adapting to diverse and complex data environments. Its modular design and specialized components offer a robust solution for addressing the challenges associated with managing heterogeneous data sources in a retrieval-augmented generation context.

\section{System Workflow}
The system workflow begins with the submission of a natural language query by the user. This query acts as the input for the system and forms the basis for subsequent operations. The query, often posed in plain language, encapsulates the user's request and is the starting point for the retrieval and response generation process.

Following the query submission, the system generates a prompt by combining the user's query with relevant database schema details and a few-shot example set. These few-shot examples demonstrate the expected query formats and outputs, helping the system better interpret and process the user query in the context of the target database. The formation of this prompt ensures that the query is accurately tailored to the database's structure and requirements.

Once the prompt is prepared, the system identifies the most appropriate query generation agent based on the type of data source the user intends to query. These agents are specialized for specific database types, such as relational, document-based, or graph databases. The selected agent generates an optimized query that adheres to the database's schema and fulfills the user's requirements. This step ensures that the generated query is both syntactically correct and semantically aligned with the data source.

The generated query is then executed within the query execution environment. This component of the system interacts with the appropriate database, ensuring seamless compatibility with diverse database systems. It retrieves the relevant data in response to the query, ensuring that the returned results are accurate and contextually appropriate.

After retrieving the data, the system combines it with the original query context to prepare it for further processing. This integration step is critical as it provides the generative agent with the necessary information to produce an accurate and coherent response. The retrieved data and the original query context together serve as inputs for the generative phase.

Finally, the generative agent synthesizes the final response. Leveraging the retrieved data and the user query, the agent produces a structured and contextually enriched output. The response is formatted according to the user's requirements, which may include text summaries, graphs, tables, or detailed reports. This output is then presented to the user in an easily interpretable format, completing the workflow.

\begin{figure}[H]
    \centering
    \includegraphics[width=0.8\textwidth]{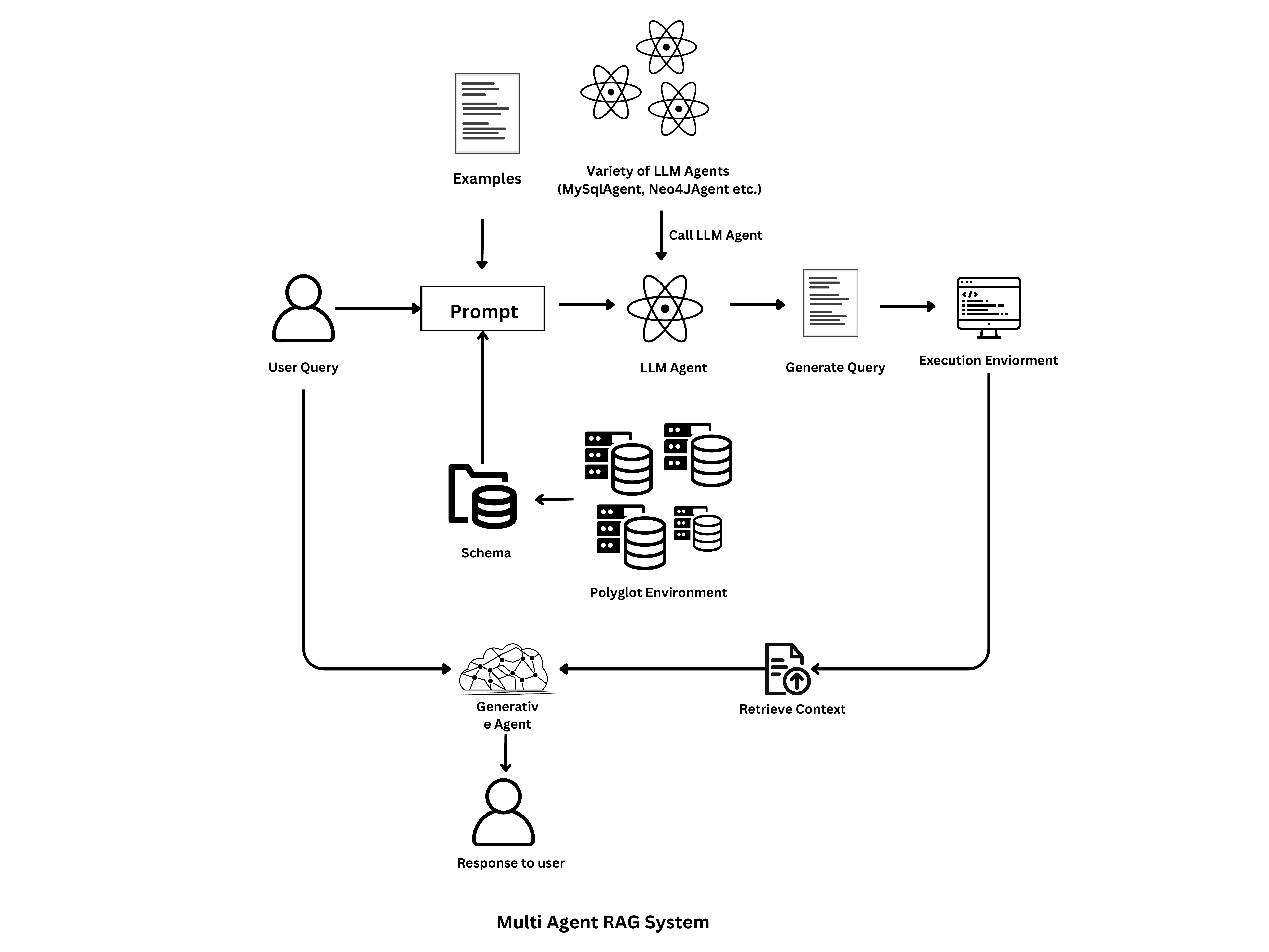}
    \caption{System Workflow of Multi-Agent RAG Agent}
    \label{fig:workflow}
\end{figure}

\section{Methodology}
The methodology adopted in the proposed system outlines the structured process through which user queries are processed to generate contextually accurate and meaningful responses. This system ensures efficiency and scalability by dividing tasks into three primary phases: query generation, query execution, and response generation. Each phase is carefully orchestrated to leverage specialized agents and a generative agent for optimal performance.

\subsection{Query Generation}
The first phase of the methodology involves processing the user’s query to identify the data source type and select the appropriate query generation agent. Each agent is equipped with the schema of the database and a few-shot prompt that guides it in generating a query tailored to the database's structure. This phase ensures that the queries are syntactically and semantically aligned with the underlying data source, providing a foundation for accurate data retrieval.



\begin{algorithm}
\caption{Query Generation Phase}
\begin{algorithmic}[1]
\State \textbf{Input}: User query $Q$, Database schema $S$
\State DataSourceType $\gets$ identify\_data\_source($Q$)
\State \textbf{Assign} Agent $\gets$ appropriate\_query\_generation\_agent(DataSourceType)

\If{Agent is available}
    \State GeneratedQuery $\gets$ Agent.generate\_query($Q$, $S$)
\Else
    \State \Return No suitable agent found for query generation.
\EndIf
\State \textbf{Output}: GeneratedQuery
\end{algorithmic}
\end{algorithm}

\subsection{Query Execution}
Once the query is generated, it is passed to the query execution environment. This environment interacts with the appropriate database drivers to execute the query and retrieve the required data. By supporting polyglot environments\cite{glake2022polyglot}, the query execution phase ensures compatibility with diverse database systems, ranging from relational databases to document stores and graph-based systems.

\begin{algorithm}
\caption{Query Execution Phase}
\begin{algorithmic}[1]
\State \textbf{Input}: GeneratedQuery, DataSourceType
\State DatabaseConnection $\gets$ connect\_to\_database(DataSourceType)
\State QueryResult $\gets$ DatabaseConnection.execute(GeneratedQuery)
\State \textbf{Output}: QueryResult
\end{algorithmic}
\end{algorithm}

\subsection{Response Generation}
The final phase combines the original user query and the retrieved data to generate a user-centric response. The generative agent synthesizes the retrieved context into a structured and meaningful output that aligns with the user’s requirements. The response can take various formats, such as plain text, tables, or visualizations like graphs, depending on the nature of the query.

\begin{algorithm}
\caption{Response Generation Phase}
\begin{algorithmic}[1]
\State \textbf{Input}: User query $Q$, QueryResult
\State Response $\gets$ GenerativeAgent.generate\_answer($Q$, QueryResult)
\State \textbf{Output}: Response
\end{algorithmic}
\end{algorithm}

\subsection{Integrated Methodology}
The overall methodology integrates the three phases into a cohesive process that transforms user queries into contextually accurate responses. The flow of the methodology ensures specialization, scalability, and efficiency, as shown in the pseudo code below.




\begin{algorithm}
\caption{Multi-Agent RAG System}
\begin{algorithmic}[1]
\State \textbf{Input}: User query $Q$
\State DataSourceType $\gets$ identify\_data\_source($Q$)
\State Agent $\gets$ appropriate\_query\_generation\_agent(DataSourceType)

\If{Agent is available}
    \State GeneratedQuery $\gets$ Agent.generate\_query($Q$, schema)
    \State QueryResult $\gets$ execute\_query(DataSourceType, GeneratedQuery)
    \State Response $\gets$ GenerativeAgent.generate\_answer($Q$, QueryResult)
\Else
    \State \Return No suitable agent found for query generation.
\EndIf

\State \textbf{Output}: Response
\end{algorithmic}
\end{algorithm}

\section{Future Scope}

The proposed Multi-Agent RAG system introduces a modular and scalable architecture, but there are several exciting directions for future research and development. Enhancing inter-agent communication and collaboration is a key area of focus. By enabling agents to share intermediate insights or partial results, the system could better handle multi-faceted and complex queries, ensuring a seamless flow of information across agents. This improvement would enable the system to address cross-domain challenges more effectively and boost its overall performance in handling intricate workflows.

Another promising avenue lies in incorporating adaptive learning mechanisms. By embedding feedback loops into the system, the generative agent and query generation agents could evolve to refine their outputs dynamically. This approach would allow the system to learn from user interactions, improving its ability to produce accurate and contextually relevant responses over time, even as the data landscape changes. Adaptive learning ensures the system remains robust and responsive to new and complex challenges.

Optimizing prompt engineering strategies is also an essential area for improvement. Fine-tuning prompts to maximize agent efficiency could lead to better query generation and response synthesis. By leveraging advanced methods for prompt design, the system could minimize token overhead, reduce latency, and ensure responses are both precise and coherent.\cite{trad2024prompt, yang2023improve}

\section{Conclusion}
The proposed Multi-Agent Retrieval-Augmented Generation (RAG) system represents a significant advancement in leveraging generative AI for diverse and complex data environments. By introducing specialized agents tailored for different database types, a centralized query execution environment, and a generative agent for synthesizing responses, this framework addresses critical limitations of traditional single-agent RAG systems. It enhances query precision, optimizes token usage, and ensures scalability across heterogeneous data sources.

The system’s modular architecture demonstrates adaptability across various industries, from healthcare to logistics, where seamless integration with relational, document-based, and graph databases is vital. Furthermore, the focus on error handling, efficient resource utilization, and reducing computational overhead establishes the proposed solution as robust and reliable for real-world applications.

Looking ahead, this research lays the foundation for future advancements in multi-agent systems, including improved inter-agent collaboration, adaptive learning capabilities, and refined prompt engineering strategies. These innovations have the potential to further enhance the system's efficiency and versatility, making it an indispensable tool for solving increasingly complex and data-intensive problems. This work not only bridges existing gaps in RAG methodologies but also opens new avenues for the thoughtful integration of AI into dynamic and diverse data ecosystems.

\bibliographystyle{unsrt}

\end{document}